\documentclass[letterpaper]{article} 
\usepackage[submission]{aaai25}  
\usepackage{times}  
\usepackage{helvet}  
\usepackage{courier}  
\usepackage[hyphens]{url}  
\usepackage{graphicx} 
\usepackage{natbib}  
\usepackage{caption} 
\usepackage{algorithm}
\usepackage{algorithmic}

\usepackage{amsmath}
\usepackage{multirow}
\usepackage{booktabs}
\usepackage{subcaption}
\usepackage{amssymb}

\usepackage{newfloat}
\usepackage{listings}
\DeclareCaptionStyle{ruled}{labelfont=normalfont,labelsep=colon,strut=off} 
\floatstyle{ruled}
\newfloat{listing}{tb}{lst}{}
\floatname{listing}{Listing}
\def\showauthors@on{T}

\title{TextToucher: Fine-Grained Text-to-Touch Generation}

\author{
    Jiahang Tu\textsuperscript{\rm 1},
    Hao Fu\textsuperscript{\rm 1},
    Fengyu Yang\textsuperscript{\rm 4},
    Hanbin Zhao\thanks{Corresponding author.  Code is available at: \protect\url{https://github.com/TtuHamg/TextToucher}}\textsuperscript{\rm 1,\rm 2},
    Chao Zhang\textsuperscript{\rm 1,\rm 2},
    Hui Qian\textsuperscript{\rm 1,\rm 3}
}
\affiliations{
    \textsuperscript{\rm 1}College of Computer Science and Technology, Zhejiang University\\
    \textsuperscript{\rm 2}Advanced Technology Institute, Zhejiang University\\
    \textsuperscript{\rm 3}State Key Lab of CAD\&CG, Zhejiang University\\
    \textsuperscript{\rm 4}College of Computer Science and Technology, Yale University\\
    \{tujiahang, haof.pizazz, zhaohanbin, zczju, qianhui\}@zju.edu.cn, fengyu.yang@yale.edu
}

\usepackage{bibentry}

\begin{document}

\maketitle

\begin{abstract}
    Tactile sensation plays a crucial role in the development of multi-modal large models and embodied intelligence. To collect tactile data with minimal cost as possible, a series of studies have attempted to generate tactile images by vision-to-touch image translation. However, compared to text modality, visual modality-driven tactile generation cannot accurately depict human tactile sensation. In this work, we analyze the characteristics of tactile images in detail from two granularities: object-level (tactile texture, tactile shape), and sensor-level (gel status). We model these granularities of information through text descriptions and propose a fine-grained Text-to-Touch generation method (TextToucher) to generate high-quality tactile samples. Specifically, we introduce a multimodal large language model to build the text sentences about object-level tactile information and employ a set of learnable text prompts to represent the sensor-level tactile information. To better guide the tactile generation process with the built text information, we fuse the dual grains of text information and explore various dual-grain text conditioning methods within the diffusion transformer architecture. Furthermore, we propose a Contrastive Text-Touch Pre-training (CTTP) metric to precisely evaluate the quality of text-driven generated tactile data. Extensive experiments demonstrate the superiority of our TextToucher method.
\end{abstract}

\section{Introduction}

Tactile sensation is one of the earliest developed senses in humans~\cite{touch_ns}. Infants begin to explore the world by touching objects, which enables them to build up their cognition of texture and shape. In the field of Multimodal Large Language Models (MLLM), researchers recognize the importance of tactile sensation in physical reasoning~\cite{octopi, newton} and tactile sensation is regarded as an important component in multimodal learning~\cite{kerr2022self, zambelli2021learning}. It also plays a fundamental role in embodied intelligence to
interact with the environments
~\cite{wu2024vision, barreiros2022haptic}.
To construct effective tactile-based multimodal models, a large volume of high-quality tactile data is required; however, obtaining tactile data is not straightforward. On the one hand, manual collection~\cite{touchandgo, hct, song2020grasping, sundaram2019learning, owens2016visually} is quite labor-intensive and time-consuming; on the other hand, robotic collection\cite{calandra2018more, li2019connecting, murali2018learning, ssvtp} limits the ways of interacting with objects, lacking flexibility and diversity. Therefore, high-quality generation of tactile data\cite{unitouch, touch_gen1, touch_gen2} is becoming a frontier research area.

\begin{figure}[t]
    \centering
    \includegraphics[width=1.0\columnwidth]{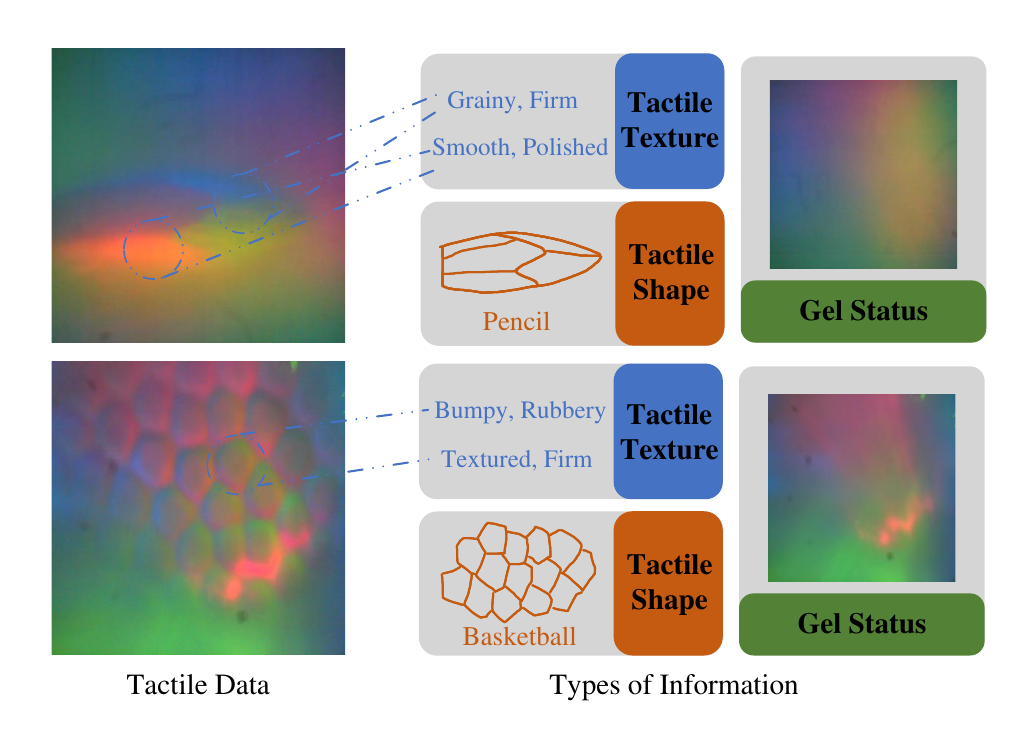}
    \caption{We present tactile images captured by sensors under different gel statuses. In our opinion, each tactile image contains three types of information: tactile texture, tactile shape, and gel status.}
    \label{fig:types_condition}
\end{figure}

In recent years, as large-scale generative models~\cite{dalle, sde} have demonstrated outstanding generative capabilities, some vision-conditioned methods attempt to utilize these models to generate tactile images by visual-to-tactile image translation ~\cite{touch_gen1, touch_gen2, unitouch}. Nevertheless, we believe these methods have two fatal shortcomings. \textbf{1)}  Humans tend to describe tactile sensation using text rather than visual images. Consequently, text-conditioned generation methods can utilize more accurate information descriptions to produce data that align closely with tactile experiences ~\cite{obrist2013talking}. \textbf{2)} Tactile images capture the deformation of objects on an elastomer gel~\cite{gel1, gel2} which is embedded with cameras and lighting systems. This implies that setting the same object on tactile sensors under different gel conditions, including changes in lighting design, camera placement, and gel material, will produce distinctly different tactile images. To our knowledge, existing works ~\cite{touch_gen1, unitouch, touch_gen2} on tactile generation have not considered the impact of different gels on the quality of the synthesized data.

As people usually describe tactile sensations in terms of the smoothness or softness, a straightforward approach is to use texture descriptions~\cite{picard2003perceptual} to guide tactile image generation. However, we carefully analyze the characteristics of tactile images and divide them into two granularities: \textbf{1) Object-level}. Tactile texture~\cite{hollins1993individual} and tactile shape~\cite{gel1} are the two types of information related to the
object in tactile images. Tactile texture is the primary attribute associated with tactile perception, such as smoothness and softness; tactile shape refers to the shape of the contact surface of the
object and is manifested through changes in color and brightness. \textbf{2) Sensor-level}. Gel status includes the sensor information about the position of cameras, light sources, and gel material. In Fig.~\ref{fig:types_condition}, we extract texture and shape information from tactile images, and present two different gel statuses, which are reflected in the tactile images collected without contact. The gel status affects the color variations of both tactile texture and shape, making this information crucial for the generation of tactile images.

In this paper, we introduce TextToucher, the first method specially designed for the text-to-touch generation task. We analyze the characteristics of tactile images in detail from two granularities: object-level (tactile texture, tactile shape), and sensor-level (gel status), which are modeled through text descriptions. For object-level conditions, we employ a multimodal large language model to build text sentences and design a question template tailored for the tactile collection situation to improve the accuracy of model responses. For sensor-level conditions, a set of learnable text prompts is defined to represent gel statuses. We further employ a pre-trained text model to encode text sentences and propose a time-adaptive strategy to fuse the tactile information. In the diffusion transformer architecture, various dual-grain text conditioning methods are explored to better control the tactile generation. Additionally, we introduce the Contrastive Text-Touch Pre-training (CTTP) metric, akin to CLIP~\cite{clip}, for evaluating the alignment between the generated tactile images and the text conditions. Extensive results demonstrate that through fine-grained textual conditions, TextToucher can effectively generate high-quality tactile images.

In summary, our contributions can be outlined as follows:
\begin{itemize}
    \item We are the first to explore the text-to-touch generation task and demonstrate text-conditioned methods are more suitable for tactile generation than vision-conditioned methods.
    \item We conduct an in-depth analysis of tactile images, identifying two granularities of tactile images: object-level (tactile texture, tactile shape) and sensor-level (gel status). With fine-grained textual conditions, TextToucher can effectively synthesize high-quality tactile images.
    \item We introduce the CTTP metric, a new measure for evaluating the alignment between generated tactile images and textual descriptions.
\end{itemize}

\section{Related Work}
\subsection{Tactile Sensor}
Recent years have witnessed the development of different tactile sensors in many robotic applications, including sliding detection, texture recognition, object pushing, insertion, and tightening. Initial tactile sensors were designed to measure force, vibration and temperature by capturing simple, low-dimensional sensory signals. Lately, vision-based tactile sensors (e.g., GelSight~\cite{gel2, gel1, johnson2011microgeometry}, GelTip~\cite{gomes2020geltip}, TacTip~\cite{ward2018tactip}, DIGIT~\cite{digit}) have been proposed
and utilize the deformation of an illuminated membrane to provide detailed information about shape and material properties. Compared to traditional single-point tactile sensors and tactile arrays, these vision-based tactile sensors offer higher-resolution tactile data. With the development of tactile sensors, a variety of tactile datasets are created by simulation-based~\cite{objectfolder1, objectfolder2, objectfolder3}, generation model-based~\cite{touch_gen1, unitouch, touch_gen2, dou2024tactile}, human-collected methods~\cite{hct, ssvtp, touchandgo}.
Our approach belongs to the generation model-based methods and focuses on generating high-resolution tactile data with diffusion generative models.

\begin{figure*}[t]
    \centering
    \includegraphics[width=0.9\textwidth]{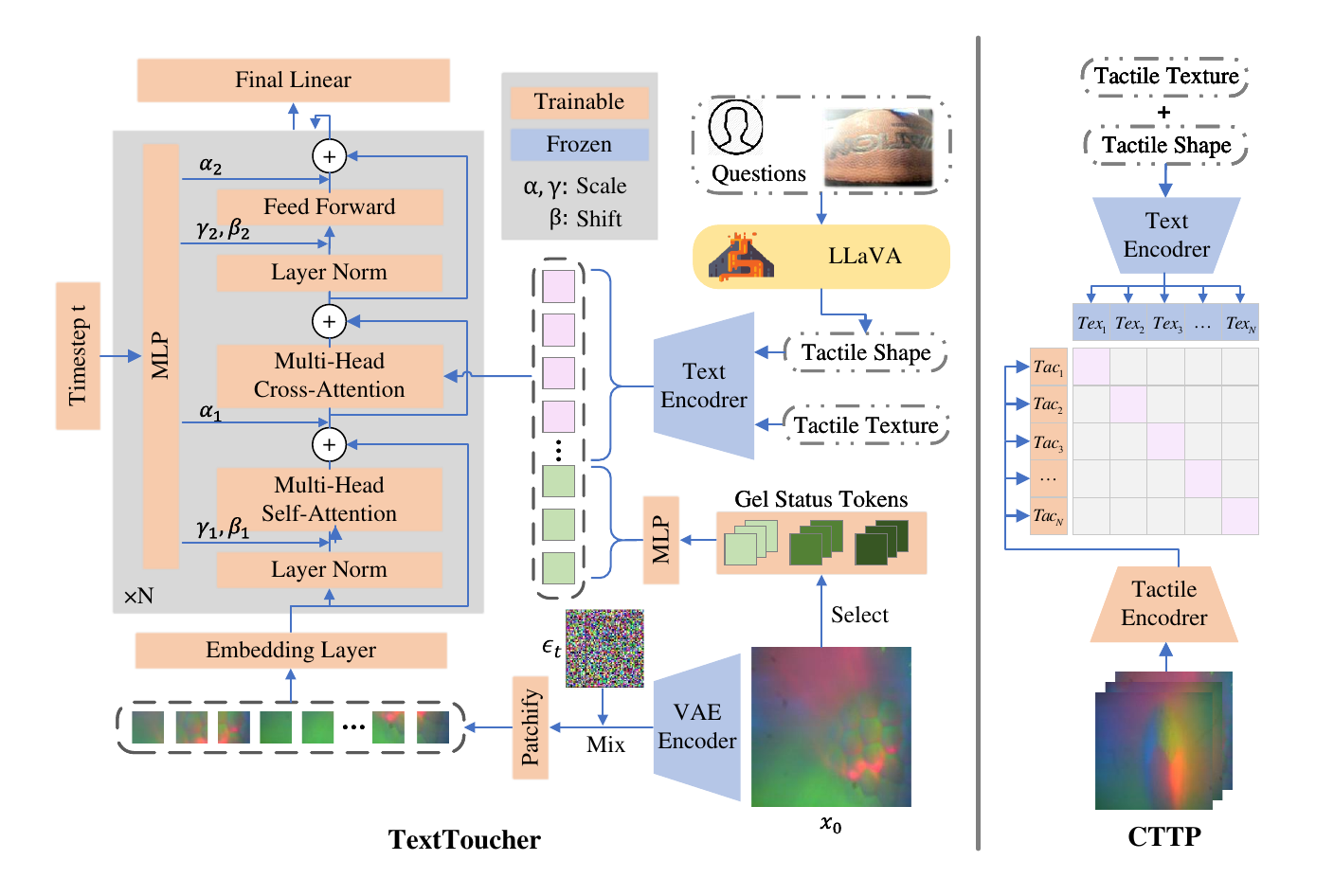}
    \caption{\textbf{Left:} Our proposed TextToucher utilizes text modality to obtain tactile texture, tactile shape and gel status information. We employ LLaVA, a vision-language large model, to caption the shape information in tactile images. Combining with texture descriptions from tactile datasets, we encode them with a text encoder. Additionally, we define a set of special word tokens to represent gel status information. \textbf{Right:} We train a tactile encoder using a contrastive loss function. In the shared space of text and tactile modalities, we propose a metric called CTTP, which uses cosine similarity to represent the relationship between tactile images and text descriptions. Our metric aims to effectively evaluate the quality of text-conditioned tactile image generation. }
    \label{architecture}
\end{figure*}

\subsection{Cross-Modal Synthesis with Generative Models}
An emerging line of work has addressed the challenges of learning from cross-modal synthesis with generative models. ImageBind~\cite{girdhar2023imagebind} bridges multi-modality within a joint embedding space, it aligns the encoders of audio, video, depth, and text with the image encoder, thereby subtly integrating all five modalities and facilitating downstream cross-modal tasks, such as tactile image generation. Building on ImageBind, ImageBind-LLM~\cite{han2023imagebind} incorporates a Large Language Model (LLM), enhancing the model’s multi-modal understanding and reasoning capabilities. ImageBind-LLM employs a visual encoder to connect LLM with all other encoders; consequently, by leveraging imageBind-LLM, the images can be generated using LLM or other modalities.

In the field of touch modality, Vision2Touch.~\cite{li2019connecting} introduces the VisGel dataset, which consists of tactile-visual paired images, and employs conditional GANs to achieve cross-modal image synthesis between GelSight tactile images and visual images. GVST~\cite{yang2022touch} proposes the Touch and Go dataset, which encompasses multiple scenarios of visuo-tactile paired images, and utilizes diffusion models to accomplish cross-modal synthesis tasks from tactile images to visual images. UniTouch~\cite{yang2024binding} also integrates the tactile modality into ImageBind-LLM, enabling the model to generate tacile images from various tactile sensors such as GelSight, DIGIT, and GelSlim\cite{taylor2022gelslim}. However, few works directly establish a bridge between text and tactile modalities, addressing the text-to-touch generation task.

\section{Methodology}
Texture, shape, and gel status are three significant attributes of tactile representations. The well-explored modality, text, with its rich semantics, is a wise choice for expressing the aforementioned attributes. In light of this, we aim to translate text to tactile images using a generative model. we utilize the Diffusion Transformer (DiT)\cite{dit} to conduct this task.

\subsection{Preliminaries}
Before introducing our method for tactile generation, we briefly review the fundamental of diffusion probabilistic models \cite{ddpm, score} (DPMs). Like most generative models\cite{vae, gan}, DPMs need to learn a mapping from a simple distribution, such as a Gaussian distribution, to the distribution of datasets. Given a data distribution $x_0\sim q_{data}(x)$, we define that the noising process iteratively adds Gaussian noise $\epsilon_t\sim \mathcal{N}(0,I)$ to the sample data $x_0$ until $x_T$. This process can be described as follows:
\begin{equation}
    q(x_t|x_{t-1})=\mathcal{N}(x_t;\sqrt{1-\beta_t}x_{t-1},\beta_t\textbf{I}), \label{eq:one_step}
\end{equation}
where $\beta_t$ denotes the noise intensity at each timestep $t$. As highlighted by DDPM~\cite{ddpm}, the Eq.~\ref{eq:one_step} can be simplified with $\alpha_t=1-\beta_t$ and $\bar\alpha_t=\prod_{s=1}^t\alpha_s$:
\begin{align}
    x_t & = \sqrt{\bar\alpha_t}x_0 + \sqrt{1-\bar\alpha_t}\epsilon, \label{eq:derive_1}
\end{align}
where $\epsilon\sim \mathcal{N}(0,\textbf{I})$. DPMs adopt a neural network to represent $p_\theta(x_{t-1}|x_t)$ and approximate the posterior $q(x_{t-1}|x_t,x_0)$ with it. Then the loss function can be written as follows:
\begin{align}
    \mathcal{L}_{\text{simple}} & = E_{t,x_0,\epsilon_t}\left[ || \epsilon - \epsilon_{\theta}(x_t, t) ||^2 \right].
\end{align}
where $\epsilon_{\theta}(x_t, t)$ is the predicted noise by diffusion model at time $t$.

To make image generation more controllable, conditional diffusion models incorporate additional inputs, such as textural descriptions\cite{sd, pixart} and segmentation map\cite{controlnet}. Classifier-free guidance\cite{cfg} aims to find a $x$ that maximizes $\log p(c|x)$. With Bayes theorem, the model $\tilde \epsilon_\theta(x_t,t,c)$ can be modified as follows:
\begin{equation}
    \tilde \epsilon_\theta(x_t,t,c) \propto s\cdot \epsilon_\theta(x_t,t,c) + (1-s) \cdot \epsilon_\theta(x_t,t,\emptyset),
\end{equation}
where $s$ represents the scale of the guidance and $c = \emptyset$ indicates DPMs generate data with non-conditions. Since classifier-free guidance can significantly improve the quality of vision images, we also adopt the technique in tactile image generation.

\subsection{Multimodal Large Language Model Annotation}
This subsection discusses how to construct the object-level conditions corresponding to a tactile image. Existing tactile datasets\cite{hct,ssvtp} already include tactile images along with texture descriptions and visual images. For simplicity, we derive the tactile shape based on these datasets.

\begin{figure}[t]
    \centering
    \includegraphics[width=0.9\linewidth]{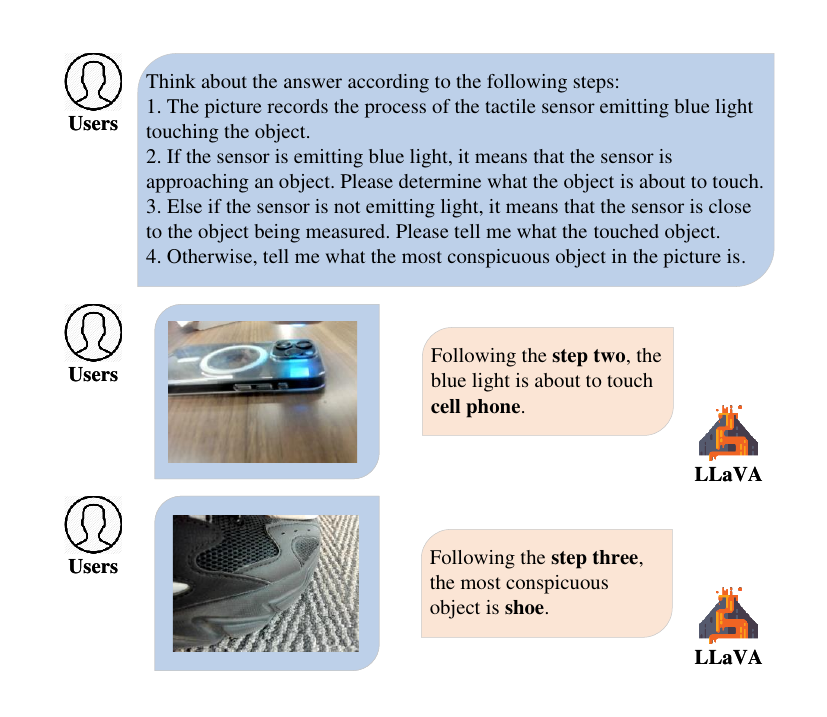}
    \caption{LLaVA engages in a step-by-step reasoning process based on carefully designed questions to achieve accurate data annotation.}
    \label{MLLM_caption}
\end{figure}

As demonstrated in Fig.~\ref{fig:types_condition}, we plan to use the text modality to describe the tactile shape. For example, a tactile image collected from the surface of a basketball is composed of interlocking pentagons; however, only a few tactile images can be precisely described the shape in words, as most shape in tactile images is irregular. TextToucher proposes to use the object that is contacted as a surrogate for tactile shape. Our motivation for this approach stems from the human ability to associate objects with their corresponding shape. When we mention a pencil, we can always imagine its shape. TextToucher aims to incorporate this ability into the generation of tactile images, allowing the model to learn the relationship between objects and their tactile shape.

Since manually labeling tactile images is time-consuming, we adopt LLaVA\cite{llava}, a large language-vision model, to caption objects in the corresponding vision images automatically.  As shown in Fig.~\ref{MLLM_caption}, we immerse the model in the process of tactile data acquisition. By explicitly providing descriptions of the stages ``about to touch'', ``in contact'', and ``special situations'', we guide the model to complete reasoning step by step\cite{zhang2022automatic}, thereby enhancing LLaVA's understanding of the acquisition scenarios. Additionally, we check the annotation results of step four to ensure the accuracy of the model's annotations.

\subsection{Gel Status Prompts}

For sensor-level conditions, the gel status encompasses the placement of cameras and light sources, as well as the material properties of gels in vision-based tactile sensors. This status is evident in the tactile images captured by the sensor when it is not in contact with any object. The sensor-level conditions significantly influence the tactile images when the sensor is in contact with objects. Therefore, in addition to object-level conditions, we consider the gel status to be a critical aspect of tactile images. To represent different gel statuses, TextToucher employs special words. Specifically, we define a set of learnable prompts $c^{sen}_i=(c^1_i, c^2_i,\dots ,c^{n_{gs}}_i)$ to denote the $i$-th gel status, where each status consists of $n_{gs}$ tokens. These prompts capture the unique characteristics of each gel status, enabling the model to accurately reflect the influence of the gel properties on the generated tactile images.

\subsection{Dual-Grain Text Conditioning Design}
The information of tactile images is divided into two granularities: the object level and the sensor level. For object-level conditions, we specially design a text template $P$ that includes both tactile texture $p_t$ and tactile shape $p_s$: ``the touch of [$p_s$] is [$p_t$]''. In the text-to-touch generation task,  T5 language model\cite{t5} is applied to translate tactile texture and shape conditions to embeddings $c^{obj}=(c_1, c_2,\dots ,c_l)\in \mathcal{R}^{l\times d_c}$.
To refine condition fusion, we propose a time-adaptive condition method. Specifically, we first concatenate dual grains of text conditions $\tilde c = (c^{sen}, c^{obj})$ to generate a rough image containing gel
information when the sampling timestep $t$ is larger than a certain threshold $\theta_t$. After that, object-level conditions are applied to focus on generating the tactile details.

We further design three different conditioning mechanisms and explore how to apply texture, shape and gel status to guide the generation of tactile images.

\subsubsection{Text Modulation.} Similar to AdaLN in DiT\cite{dit, pixart}, we explore modulating the output of layer norm layers, self-attention layers and MLP layers. Rather than directly translate class embeddings to scale and shift parameters $\gamma$ and $\beta$, we employ learnable MLP layers to fuse the sequence information in condition $\tilde c$ and add the result to timestep $t$.
\subsubsection{Joint Attention.} We concatenate the text tokens $\tilde c$ with the input tokens $x \in \mathcal{R}^{n\times d_c}$ of transformer blocks. The joint sequence  $\tilde x = [c^{sen}, c^{obj}, x_1, x_2,\dots, x_n]$ is then fed into the self-attention layer. To ensure the scalability of  transformer blocks, the output discards tokens related to the text conditions before being fed into subsequent layers.
\subsubsection{Cross Attention.} We insert a multi-head cross attention layer after the multi-head self-attention layer within the transformer block. Each tactile token calculates the attention scores with text tokens that include texture, shape, and gel status. As shown in Fig.~\ref{architecture}, we retain the modulation part of timestep embeddings $t$ as employed in DiT, which is crucial for incorporating temporal information into the model.

\subsection{Contrastive Text-Touch Pre-training (CTTP)}

In the text-to-touch generation task, it is difficult to evaluate the correlation between a generated tactile image and its text description by visual inspection. Following other works in cross-modal synthesis\cite{touch_gen1}, we propose a new metric, Contrastive Text-Touch Pre-training (CTTP), to measure the alignment between the tactile images and text descriptions.

We refer to the instances from the same tactile-textual record $\{tac_i, tex_i\}$ as positive pairs, and instances from different tactile-textual record $\{tac_i, tex_j\}$ as negative pairs. Our goal is to minimize the embedding distance between positive sample pairs and maximize the embedding distance between negative sample pairs. Inspired by the CLIP\cite{clip} training method, we use InfoNCE\cite{infoNCE} to maximize the probability of positive sample pairs in each training batch $B$ in Fig.~\ref{architecture}:
\begin{align}
    \mathcal{L}_i^{tac,tex} = -\log \frac{exp(E_{tac}(tac_i)\cdot E_{tex}(tex_i)/\tau)}{\sum_{j=1}^B exp(E_{tac}(tac_i)\cdot E_{tex}(tex_j)/\tau)},
\end{align}
where $E_{tac}$ and $E_{tex}$ are corresponding encoders, and $\tau$ is a temperature parameter. Similarly, we obtain the symmetric objective $\mathcal{L}_i^{tex,tac}$ and minimize it:
\begin{align}
    \mathcal{L}=\mathcal{L}^{tac,tex}+\mathcal{L}^{tex,tac}.
\end{align}
Once the tactile encoder $E_{tac}$ and text encoder $E_{tex}$ have been trained, we use Eq.~\ref{eq:cttp} to evaluate the similarity between the tactile images and text prompts:
\begin{align}
    CTTP(tac_i, tex_i) = \frac{E_{tac}(tac_i) \cdot E_{tex}(tex_i)}{\|E_{tac}(tac_i) \|_2 \cdot \| E_{tex}(tex_i)\|_2}. \label{eq:cttp}
\end{align}

\section{Experiment}
\subsection{Experiment Settings}
\subsubsection{Datasets}
We conduct experiments on two representative datasets. HCT\cite{hct} comprises visual-tactile data collected using a handheld 3D-printed data collection device. This dataset includes 43741 pairs of in-contact frames with tactile texture descriptions. Each data record contains the process of approaching, touching, sliding, and withdrawing from an object using the tactile sensor. Another dataset, SSVTP\cite{ssvtp}, utilizes a UR5 robotic arm equipped with an RGB camera and a tactile sensor to collect data from various deformable surface environments, such as clothing seams, buttons, and zippers.
\subsubsection{Metrics}
We adopt CTTP, LPIPS, SSIM and PSNR metrics to quantitatively evaluate the generated results. CTTP is employed to assess how well the tactile images align with the tactile descriptions, ensuring that the generated images accurately reflect the tactile sensations. LPIPS\cite{lpips} is employed to assess the similarity between the generated samples and target samples at the feature level. Following GVSR\cite{touch_gen1}, we use SSIM and PSNR to evaluate the consistency at the pixel level.

\begin{table}[t]
    \setlength{\tabcolsep}{1.0pt}

    \centering
    \resizebox{1.0\linewidth}{!}{
        \begin{tabular}{@{}cccccccccccc@{}}
            \toprule
            \multirow{2}{*}{Method} & \multirow{2}{*}{Reference} & \multirow{2}{*}{Type} & \multicolumn{4}{c}{HCT} &                    & \multicolumn{4}{c}{SSVTP}                                                                                                 \\ \cmidrule(l){4-7}  \cmidrule(l){9-12}
                                    &                            &                       & CTTP $\uparrow$         & LPIPS $\downarrow$ & SSIM $\uparrow$           & PSNR $\uparrow$ &  & CTTP $\uparrow$ & LPIPS $\downarrow$ & SSIM $\uparrow$ & PSNR $\uparrow$ \\ \midrule
            GVST                    & ICCV23                     & Vision2Touch          & -                       & 0.573              & 0.881                     & 19.45           &  & -               & 0.502              & 0.918           & 21.15           \\
            UniTouch                & CVPR24                     & Vision2Touch          & 0.156                   & 0.528              & 0.902                     & 19.84           &  & 0.127           & 0.555              & 0.824           & 12.42           \\
            PixArt-$\alpha$         & ICLR24                     & Text2Vision           & 0.198                   & 0.504              & 0.876                     & 20.26           &  & 0.125           & 0.497              & 0.916           & 22.35           \\
            \textbf{TextToucher}    &                            & Text2Touch            & \textbf{0.261}          & \textbf{0.427}     & \textbf{0.904}            & \textbf{22.70}  &  & \textbf{0.152}  & \textbf{0.465}     & \textbf{0.930}  & \textbf{22.43}  \\ \bottomrule
        \end{tabular}}
    \caption{Quantitative results on the tactile generation task are presented. Our method can achieve the best results on each metric.}
    \label{tab:sota}
\end{table}

\subsection{Main Results}
We categorize the comparison methods into two groups: 1) Vision-conditioned methods, where GVST~\cite{touch_gen1} and UniTouch~\cite{unitouch} utilize diffusion models with a Unet architecture to generate tactile images from visual scenes. 2) Text-conditioned methods, where PixArt-$\alpha$~\cite{pixart} translates text descriptions into visual images. We employ PixArt-$\alpha$ to complete the text-to-touch generation task, as both tasks involve processing text-based conditions.

\begin{figure*}[t]
    \centering
    \includegraphics[width=1.0\textwidth]{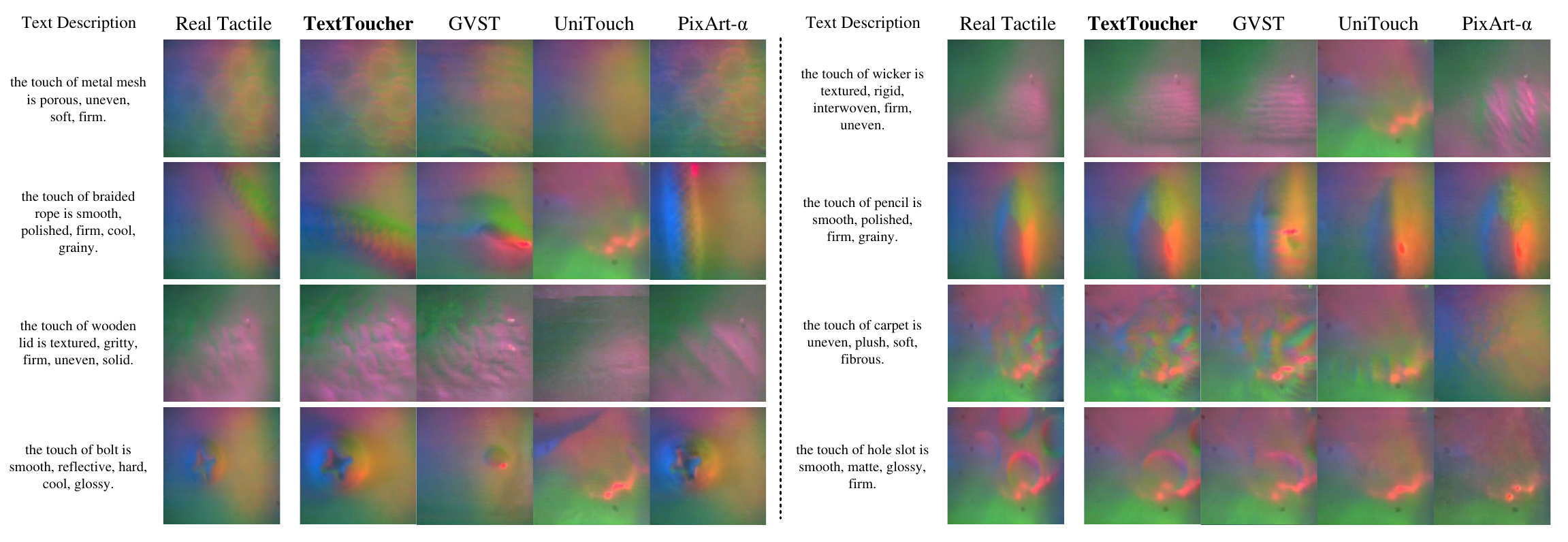}
    \caption{We compare our approach with other representative methods. TextToucher can produce tactile images with fewer artifacts and higher quality, closely aligning with the provided text descriptions.}
    \label{fig:comparison}
\end{figure*}

\begin{table*}[!t]
    \begin{subtable}[t]{0.48\textwidth}
        \centering
        \resizebox{1.0\textwidth}{!}{
            \begin{tabular}{@{}cccccccc@{}}
                \toprule
                \multirow{2}{*}{Method} & \multicolumn{3}{c}{Text Conditions} & \multirow{2}{*}{CTTP $\uparrow$ } & \multirow{2}{*}{LPIPS $\downarrow$} & \multirow{2}{*}{SSIM $\uparrow$} & \multirow{2}{*}{PSNR $\uparrow$}                                   \\ \cmidrule(r){2-4}
                                        & Tactile Texture                     & Tactile Shape                     & Gel Status                          &                                  &                                  &                &                \\ \midrule
                TC-T                    & \checkmark                          &                                   &                                     & 0.198                            & 0.504                            & 0.876          & 20.26          \\
                TC-TS                   & \checkmark                          & \checkmark                        &                                     & 0.236                            & 0.445                            & 0.887          & 22.55          \\
                TextToucher             & \checkmark                          & \checkmark                        & \checkmark                          & \textbf{0.261}                   & \textbf{0.427}                   & \textbf{0.904} & \textbf{22.70} \\ \bottomrule
            \end{tabular}
        }
        \subcaption{Comparison of different text condition types.}
        \label{ablation:conditions}
    \end{subtable}
    \hspace{0.02\textwidth}
    \begin{subtable}[t]{0.48\textwidth}
        \centering
        \resizebox{0.85\textwidth}{!}{
            \begin{tabular}{@{}ccccc@{}}
                \toprule
                Conditioning Mechanism & CTTP $\uparrow$ & LPIPS $\downarrow$ & SSIM $\uparrow$ & PSNR $\uparrow$ \\ \midrule
                Text Modulation        & 0.151           & 0.476              & 0.899           & 21.91           \\
                Joint Attention        & 0.213           & 0.453              & 0.898           & 22.67           \\
                Cross Attention        & \textbf{0.261}  & \textbf{0.427}     & \textbf{0.904}  & \textbf{22.70}  \\ \bottomrule
            \end{tabular}
        }
        \subcaption{Comparison of different conditioning mechanisms.}
        \label{ablation:mechanism}
    \end{subtable}

    \begin{subtable}[t]{0.31\textwidth}
        \resizebox{1.0\textwidth}{!}{
            \begin{tabular}{@{}ccccc@{}}
                \toprule
                Layer & CTTP $\uparrow$ & LPIPS $\downarrow$ & SSIM $\uparrow$ & PSNR $\uparrow$ \\ \midrule
                1-14  & 0.258           & 0.431              & 0.900           & 22.14           \\
                1-28  & \textbf{0.261}  & \textbf{0.427}     & \textbf{0.904}  & \textbf{22.70}  \\
                14-28 & 0.241           & 0.460              & 0.893           & 21.74           \\
                7-28  & 0.249           & 0.458              & 0.902           & 21.84           \\ \bottomrule
            \end{tabular}
        }
        \subcaption{Comparison of gel status prompts inserted in different layers.}
        \label{ablation:layers}
    \end{subtable}
    \hspace{0.02\textwidth}
    \begin{subtable}[t]{0.31\textwidth}
        \resizebox{1.0\textwidth}{!}{
            \begin{tabular}{@{}ccccc@{}}
                \toprule
                $n_{gs}$ & CTTP $\uparrow$ & LPIPS $\downarrow$ & SSIM $\uparrow$ & PSNR $\uparrow$ \\ \midrule
                1        & 0.248           & 0.452              & 0.900           & 21.74           \\
                2        & 0.255           & 0.452              & 0.895           & 21.93           \\
                4        & \textbf{0.261}  & \textbf{0.427}     & \textbf{0.904}  & \textbf{22.70}  \\
                6        & 0.250           & 0.458              & 0.899           & 21.56           \\
                8        & 0.252           & 0.453              & 0.899           & 21.60           \\\bottomrule
            \end{tabular}
        }
        \subcaption{Comparisons of token counts representing the gel status prompts.}
        \label{ablation:token_num}
    \end{subtable}
    \hspace{0.02\textwidth}
    \begin{subtable}[t]{0.31\textwidth}

        \resizebox{1.0\textwidth}{!}{
            \begin{tabular}{@{}ccccc@{}}
                \toprule
                $\theta_t$ & CTTP $\uparrow$ & LPIPS $\downarrow$ & SSIM $\uparrow$ & PSNR $\uparrow$ \\ \midrule
                800        & 0.258           & \textbf{0.427}     & 0.903           & 22.69           \\
                600        & \textbf{0.261}  & \textbf{0.427}     & \textbf{0.904}  & \textbf{22.70}  \\
                400        & 0.247           & 0.434              & 0.897           & 22.54           \\
                200        & 0.235           & 0.435              & 0.892           & 22.26           \\
                0          & 0.234           & 0.434              & 0.899           & 22.12           \\ \bottomrule
            \end{tabular}
        }
        \subcaption{The Effect of timestep threshold $\theta_t$ employing different tactile text conditions.}
        \label{ablation:timestep}
    \end{subtable}
    \caption{Ablation studies on HCT dataset with our proposed method.}
\end{table*}

\subsubsection{Quantitative Evaluation.}
The quantitative results are presented in Tab.~\ref{tab:sota}. TextToucher consistently achieves superior performance across all metrics on both HCT and SSVTP datasets. It is observed that TextToucher significantly outperforms GVST and UniTouch, confirming our hypothesis that the text modality can more accurately describe tactile sensations compared to the vision modality. Additionally, our method distinctly surpasses PixArt-$\alpha$, demonstrating a +0.063 improvement in CTTP and a +0.077 enhancement in LPIPS on the HCT dataset. It also performs comparably on the SSVTP dataset, suggesting that traditional text-to-image methods are inadequate for tactile generation tasks.

\subsubsection{Qualitative Evaluation.}
Fig.~\ref{fig:comparison} shows the qualitative comparisons with alternative methods. Provided with text descriptions, our method can generate results that are more consistent with the reference tactile images. In contrast to vision-conditioned methods, TextToucher can effectively captures the contours of the contact objects through the tactile shape conditions. Furthermore, we observe that our method produces fewer artifacts in the generated images compared to PixArt-$\alpha$. This improvement can be attributed to the dual-grain text conditions, which accurately model the intrinsic characteristics of tactile images, thereby enhancing the overall fidelity of tactile image synthesis.

\subsection{Ablation Studies}
In this section, we conduct several ablation studies to analyze the performance of our proposed method.
\subsubsection{Text Condition Types.} In Tab.~\ref{ablation:conditions}, we explore various combinations of text condition types in the text-to-touch generation task. Compared to employing only the tactile texture condition, generating tactile images with additional descriptions of tactile shapes and gel statuses contributes to more accurately forming the contours of contact objects and presenting them on specific gels. This approach achieves significant improvements across four metrics, underscoring the effectiveness of fine-grained text conditions in enhancing the quality of tactile image generation.

\subsubsection{Conditioning mechanisms.} In Tab.~\ref{ablation:mechanism}, we observe the Cross Attention conditioning mechanism is more effective for the text-to-touch generation task. In the Text Modulation approach, compressing sequential text tokens into inputs for modulation leads to a substantial loss of information. Besides, the imbalance in the number of tactile (1024 tokens) and text tokens (17.3 tokens on average) in the Joint Attention approach hampers the model's ability to establish connections between the two modalities.

\subsubsection{Gel status prompts in different layers.}
We investigate the effects of applying gel status prompts to different layers of the model via two designs: gradually adding them from shallow to deep and from deep to shallow. The results in Tab. ~\ref{ablation:layers} show that employing gel status prompts within the first 14 layers enhances the quality of generated tactile images, and the best results are achieved when it is applied up to 28 layers. Adding gel status prompts from deeper to shallower layers can degrade the model's performance. This indicates that gel state encoding is crucial for shallow layer image representation and the optimal setting is adopted as the final setting for our approach.

\subsubsection{Token number $n_{gs}$.}
We study the impact of varying the number of tokens $n_{gs}$ in gel status embeddings. As shown in Tab.~\ref{ablation:token_num}, we find that using $n_{gs}=4$ tokens effectively represents different gel status and surpasses other settings across various metrics by a margin. We hypothesize that fewer tokens ($n_{gs}=1, 2$) are insufficient to capture the gel information, while a larger number of tokens ($n_{gs}=6, 8, 10$) introduces redundant information, preventing the model from allocating weights in the cross attention layers.

\begin{figure}[t]
    \centering
    \includegraphics[width=0.90\linewidth]{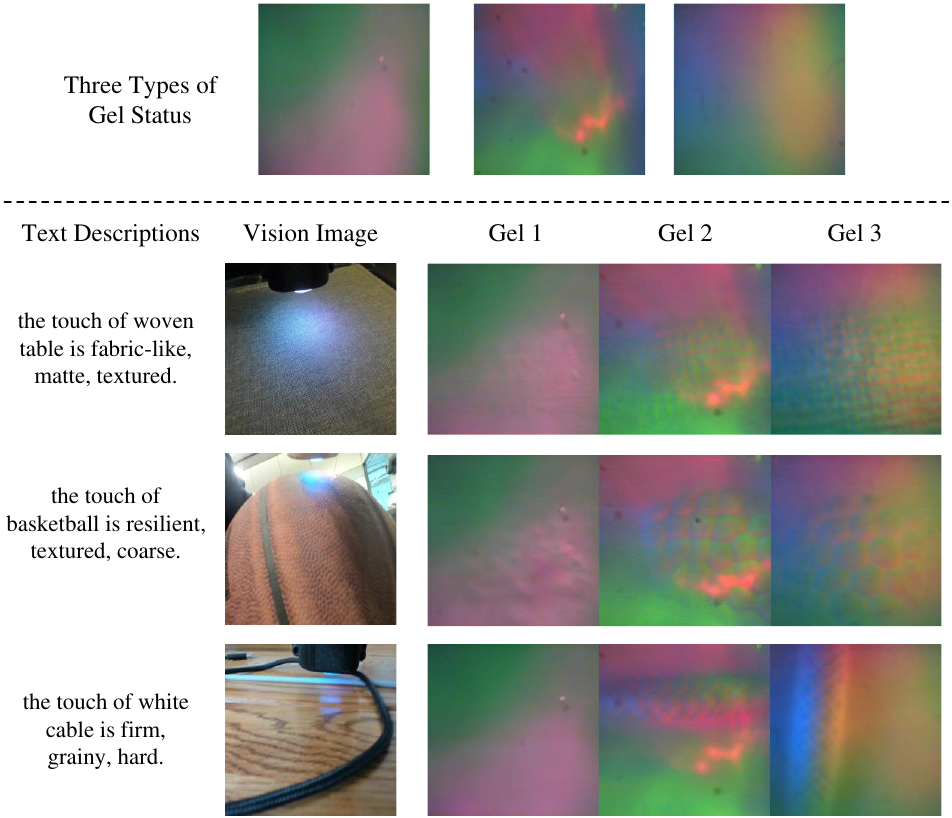}
    \caption{The first row displays the gel statuses contained in HCT dataset. We generate the same object under different gel statuses in the remaining rows.}
    \label{fig:variation_gel}
\end{figure}

\subsubsection{Timestep threshold $\theta_t$.}
We explore the impact of using different tactile text conditions at various sampling timesteps on tactile generation. We set $\tilde c = \{c^{sen}, c^{obj}\}$, which includes three types of tactile information, to be used before $\theta_t$, and the tactile texture and shape description $c$ to be used afterwards. The results in Tab.~\ref{ablation:timestep} show that this approach with time-varying tactile text conditions improves model performance, achieving the best results at $\theta_t$ of 600 or 800 timesteps. This improvement occurs because using $\tilde{c}$ in the early sampling steps helps form an initial tactile image with gel information. Subsequently, the diffusion model with the conditional encoding $c$ is able to refine the generation of tactile texture and shape.

\subsection{Variation in Gel Statuses}
In this section, we utilize different gel states to control the tactile generation conditioned on the same text sentences, enhancing the diversity of tactile images. We use three sets of prompts to represent the three gel statuses in the HCT dataset. Specifically, we use the gel state prompts to control the generation of tactile images in inference stage. The results in Fig.~\ref{fig:variation_gel} indicate that the gel status conditions effectively control the presentation of tactile shape and texture. However, we also observe that the quality of images generated for gel state 1 is poor. We analyze the dataset and find that only 2.4\% of the tactile images are collected using gel state 1. This limited sample size likely makes it difficult for the model to understand the relationship between gel state 1 and the corresponding texture and shape.

\subsection{Tactile Prediction}
\begin{figure}[t]
    \centering
    \includegraphics[width=1.0\linewidth]{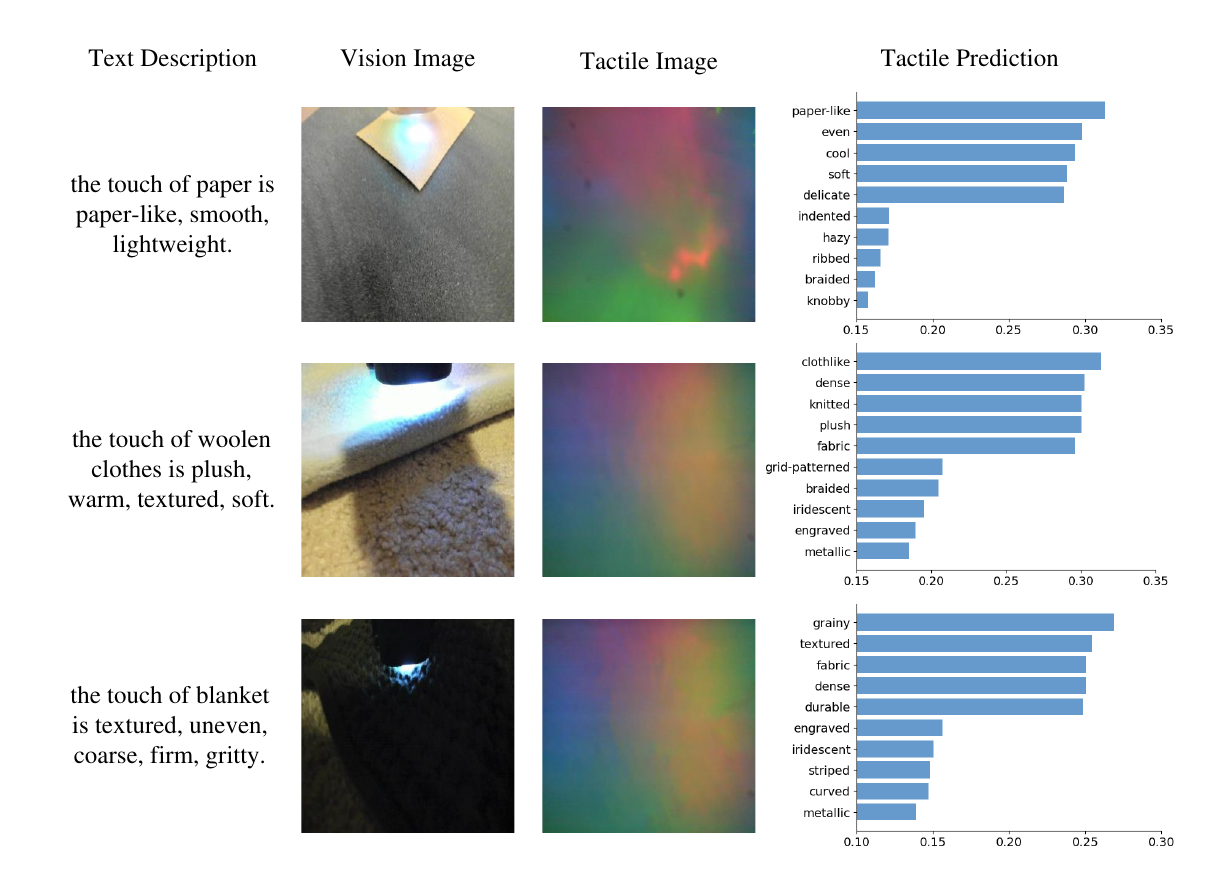}
    \caption{With the pre-trained tactile encoder, we extract features from tactile images and calculate the CTTP with all texture descriptions, predicting the top five most likely tactile texture descriptions and the most irrelevant ones.}
    \label{fig:tactile_pred}
\end{figure}

Similar to how CLIP is used for image classification, we employ the trained tactile image encoder for tactile predictions to verify the validity of CTTP metric. We extract all adjectives describing tactile texture from the HCT datasets. Using the tactile image encoder alongside the CLIP text encoder, we can compute the CTTP metric between the tactile images and all tactile descriptions $P_i = (p_s, p_{t_i})$. In Fig.~\ref{fig:tactile_pred}, we present the prediction results, including the top five highest-scoring tactile texture descriptions and the most irrelevant descriptions. For example, in the second row, terms like ``clothlike'', ``dense'', and ``knitted'' accurately describe woolen clothes, whereas ``metallic'' does not relate to clothes at all. The experimental results demonstrate that the tactile image encoder trained with contrastive learning effectively aligns with the text encoder.

\section{Conclusion}
In this paper, we are the first to propose the text-to-touch generation task and specifically analyze the characteristics of tactile images  from two granularities: object-level (tactile texture, tactile shape), and sensor-level (gel status). We propose TextToucher, a method that effectively utilizes tactile text descriptions to generate high-quality tactile images. We extend the tactile datasets containing texture descriptions and employ LLaVA to annotate the shape information of tactile images. For gel status, we define a set of learnable prompts to present different gel statuses, which can be selectively added to the text conditions based on sampling timestep. Three text conditioning mechanisms are explored to guide the generation of tactile images. Additionally, we introduce the CTTP metric to assess the alignment between tactile images and their corresponding text conditions. Experimental results and the extensive visualizations demonstrate that our method outperforms other methods and effectively generates high-quality tactile images.

\bibliography{aaai25}

\clearpage
\appendix
\onecolumn

\section{Overview}
In this supplementary material, we submit the source code in the ``TextToucher'' folder and provide more experiments about our method. The supplementary material includes: 1) More implementation details. 3) Pseudocode description of TextToucher 3) More qualitative results.

\section{Implementation Details}
\subsection{Experiment Setting}
We adopt DiT as the base architecture and select LLaVA-v1.6-34b to caption visual images in datasets, forming texture and shape descriptions. For encoding text conditions, we follow the methods of Imagen\cite{imagen} and Pixart-$\alpha$\cite{pixart}, utilizing the 4.3B Flan-T5-XXL as the text encoder. For the CTTP metric, we employ a frozen and pre-trained CLIP text encoder to obtain text embeddings and train a ViT-tiny model with 200 epochs on one RTX 4090. We use the full memory capacity of four 4090 (24G) GPUs to train our model for 200 epochs. We use the AdamW optimizer\cite{adamw} with a weight decay of 0.03 and a constant learning rate of 2e-5 after 1000 warmup steps. Gradient clipping is applied when gradients exceed 0.01 to maintain training stability. For evaluation, we use dpm-solver\cite{dpmsolver} with 50 sampling steps to generate tactile images. All experiments are conducted on Ubuntu 22.04 system with CUDA 12.2 and PyTorch 2.1.2 environment. Note that to control for randomness, the seed for all experiments, including both training and inference phases, will be set to 43, and a single evaluation will be conducted for each experiment.

\subsection{Dual-Grain Text Conditioning Methods}
\begin{figure}[H]
    \centering
    \includegraphics[width=0.95\linewidth]{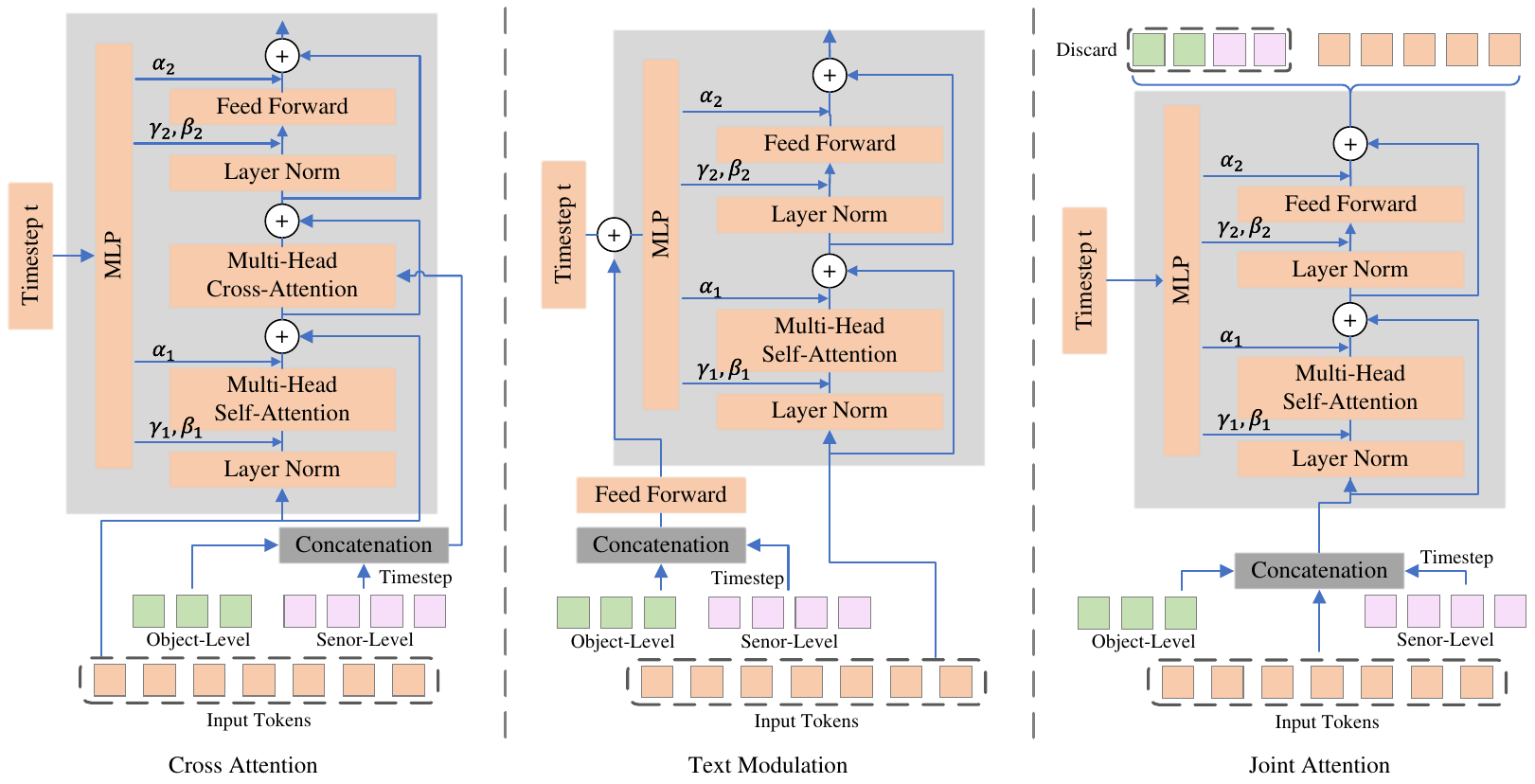}
    \caption{Different dual-grain text conditioning method.}
    \label{fig:condition}
\end{figure}

We further demonstrate different dual-grain conditioning mechanisms in Fig.~\ref{fig:condition}. We preserve the modulation component of the timestep embeddings $t$, as utilized in the DiT framework. This aspect is essential for integrating temporal information into the model, enhancing its ability to process sequences over time.

For the cross attention method, we integrate a multi-head cross attention layer subsequent to the multi-head self-attention layer. We concatenate object-level and sensor-level tokens to serve as input for the key and value in the attention layers. Each tactile token computes attention scores with text tokens that contain attributes such as texture, shape, and gel status.

For the text modulation method, we employ text conditions to modulate the outputs from layer norm layers, self-attention layers, and MLP layers. Due to a dimensionality mismatch between the token sequence and the tactile tokens, we utilize learnable MLP layers to fuse the sequence information in dual-grain text conditions. The fused output is then added to the timestep embedding, collectively influencing the generation of tactile images.

For the joint attention method, we concatenate the dual-grain tokens $\tilde{c}$ with the input tokens $x \in \mathcal{R}^{n\times d_c}$ of the transformer blocks, forming the joint sequence $\tilde x = [\tilde c, x_1, x_2, \dots, x_n]$. This sequence is then processed through the self-attention layer. To maintain consistency in the dimensions of module inputs and outputs, we discard tokens associated with the text conditions from the output prior to passing it into subsequent layers.

\subsection{Hyper-Parameters in TextToucher}
We provide additional hyper-parameters setting of our proposed TextToucher and the tactile encoder in Tab.~\ref{tab:hyperparameter}
\begin{table*}[h]
    \centering
    \begin{tabular}{@{}ll|ll@{}}
        \toprule
        Hyperparamter           & Value                 & Hyperparamter         & Value       \\ \midrule
        Learning Rate           & $2 \times 10^{-5}$    & LDM Model             & Transformer \\
        Image Size              & 512                   & LDM Input Size        & 64          \\
        Channel                 & 3                     & LDM Input Channel     & 4           \\
        Conditioning Key        & Cross Attention       & LDM Output Channel    & 8           \\
        First Stage Model       & AutoencoderKL         & LDM Channels          & 1152        \\
        VAE In-channel          & 3                     & LDM Patch Size        & 2           \\
        VAE Out-channel         & 3                     & LDM Attention Head    & 16          \\
        VAE Num. Resblocks      & 4                     & LDM Transformer Depth & 28          \\
        VAE channels            & {[}125,256,512,512{]} & Loss Type             & simple mse  \\
        Condition Model         & T5-XXL                & Diffusion Timesteps   & 1000        \\
        Condition Layer         & 24                    & Scheduler             & DPM-Solver  \\
        Condition Channel       & 4096                  & Tactile Encoder       & ViT         \\
        Condition Token Number  & 120                   & Tactile Input Size    & 224         \\
        Batch Size              & 32                    & Tactile Path Size     & 16          \\
        Epoch                   & 200                   &                       &             \\
        Training/Inference Seed & 43                    &                       &             \\ \bottomrule
    \end{tabular}
    \caption{We show detailed hyper-parameters setting of our models, including first stage model, LDM model and tactile encoder.}
    \label{tab:hyperparameter}
\end{table*}

\section{Pseudocode Description of TextToucher}
\begin{algorithm}[H]
    \begin{minipage}{\linewidth}
        \caption{TextToucher: Fine-Grained Text-to-Touch Generation}
        \begin{algorithmic}
            \STATE \textbf{Input:} dataset $\mathcal{D}$, multimodal large language model $L(\cdot, \cdot)$, question template $Q$, gel status prompts $\{c_{gs_i}\}_{i=1}^{n}$, diffusion step $T$, noise schedule $\alpha(t)$, $\sigma(t)$, Language Encoder $E_l$, Tactile Encoder $E_t$, timestep threshold $\theta_t$, diffusion model $\epsilon^c_\theta$ with different conditioning mechanism $c$
            \STATE \textbf{Stage1:}
            \FOR{each $vis_i$, $tac_i$, $tex_i^t$ in $\mathcal{D}$}
            \IF {$tac_i$ not contact}
            \STATE continue
            \ELSE
            \STATE $tex_i^s$ $\leftarrow$ $L(Q, vis_i)$, $z_i \leftarrow E_t(tac_i)$, $c_i^{obj}\leftarrow E_l(tex_i^t, tex_i^s)$
            \ENDIF
            \ENDFOR
            \STATE Organize training data into latent space: $\mathcal{D}_z=(z, c^{obj}, c^{sen})$
            \STATE \textbf{Stage2:}
            \REPEAT
            \STATE Sample $(z, c^{obj}, c^{sen}) \sim \mathcal{D}_z$, $t \sim \mathcal{U}[0,T]$
            \STATE Sample $z_t \sim \mathcal{N}(\alpha(t)z; \sigma^2(t)\textbf{I})$
            \IF{$t< \theta_t$}
            \STATE $\mathcal{L} \leftarrow \left[ || \epsilon - \epsilon^c_{\theta}(z_t, t, c^{obj}) ||^2 \right].$
            \ELSE
            \STATE $\mathcal{L} \leftarrow \left[ || \epsilon - \epsilon^c_{\theta}(z_t, t, c^{obj}, c^{sen}) ||^2 \right].$
            \ENDIF
            \UNTIL convergence
        \end{algorithmic}
    \end{minipage}
\end{algorithm}

\newpage
\section{More Qualitative Results}
\begin{figure}[H]
    \centering
    \includegraphics[width=1.0\linewidth]{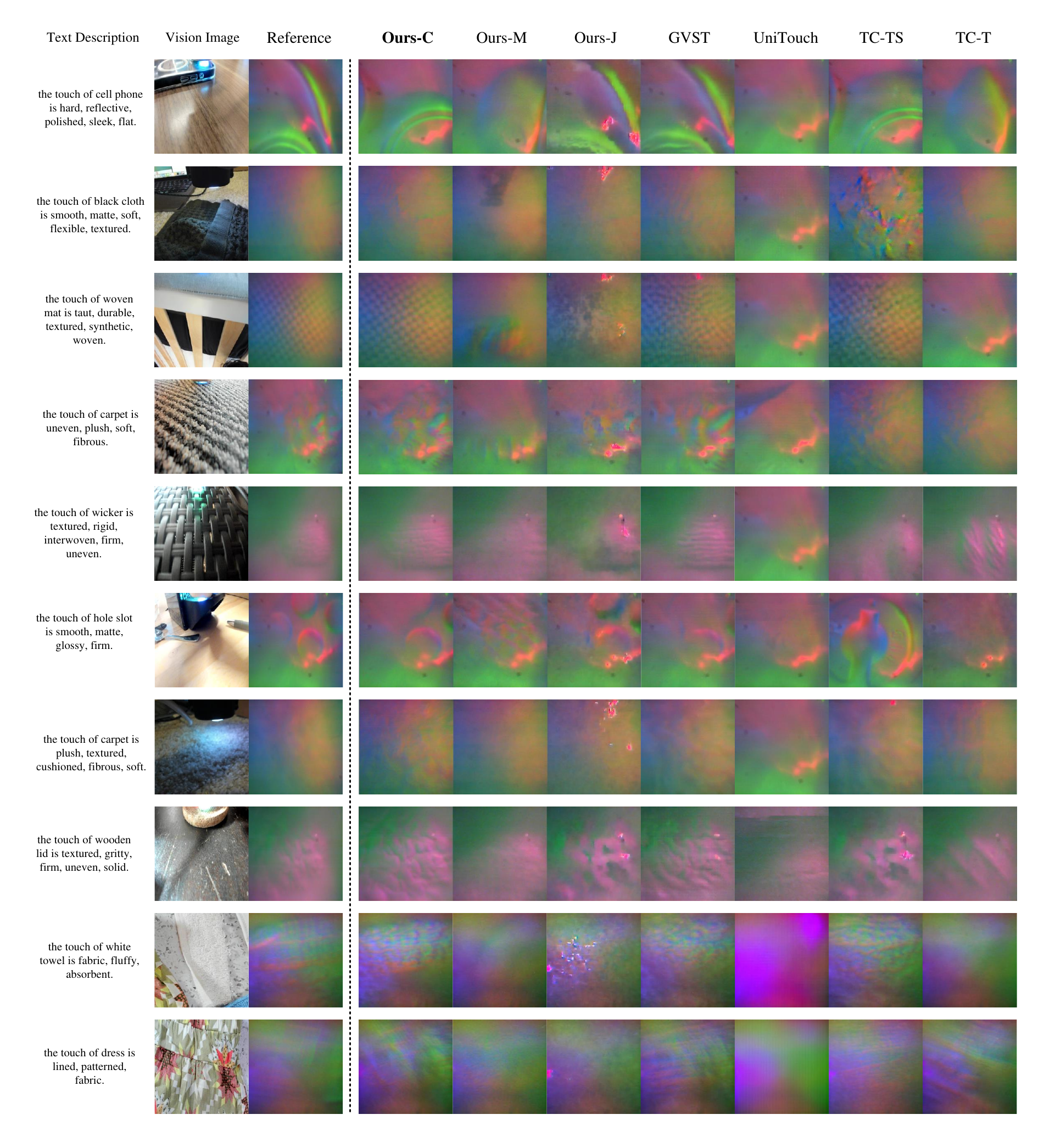}
    \caption{More comparison results with other methods. Note that we use “M”,“J”, and “C” to denote the conditions of Text Modulation, Joint Attention, and Cross Attention mechanisms, respectively.}
    \label{fig:comparison_more}
\end{figure}

\begin{figure}[H]
    \centering
    \includegraphics[width=1.0\linewidth]{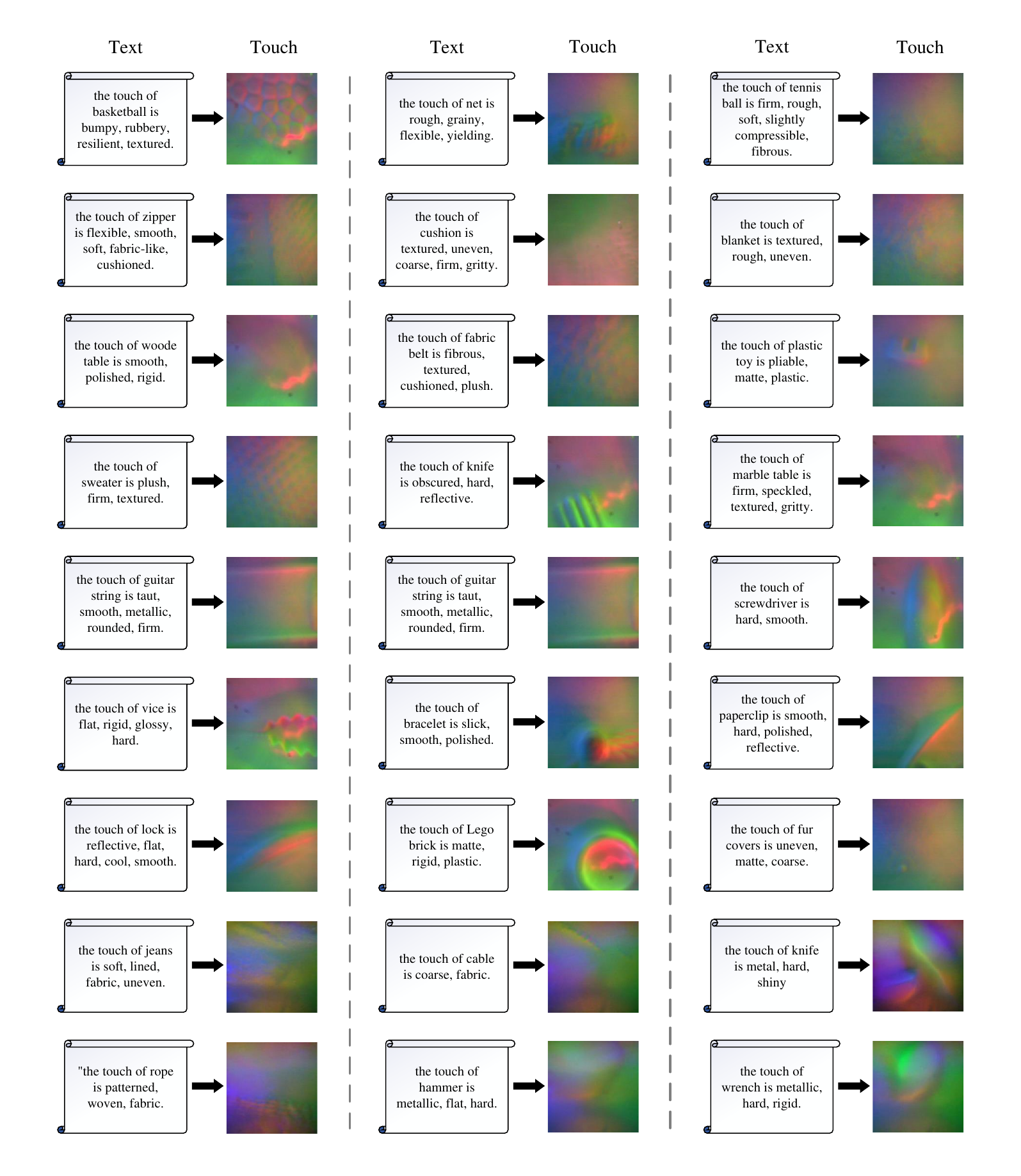}
    \caption{More generated tactile results.}
    \label{fig:generated_results}
\end{figure}

\begin{figure}[H]
    \centering
    \includegraphics[width=0.85\linewidth]{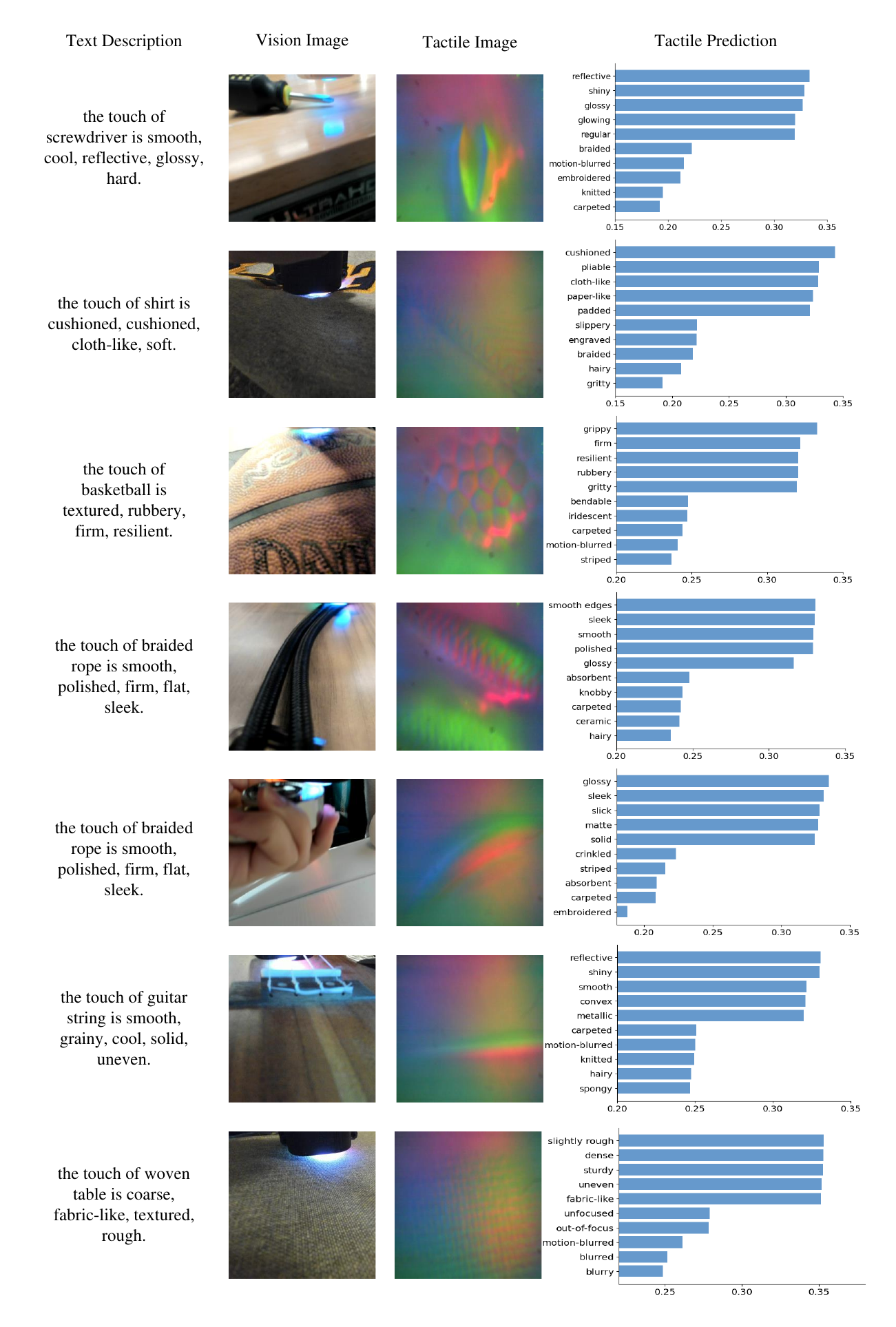}
    \caption{More tactile prediction results.}
    \label{fig:pred_results_more}
\end{figure}

In Fig.~\ref{fig:comparison_more}, we provide more comparison results between our method and other methods. Using a cross attention mechanism for dual-granularity conditioning can produce high-quality results that are more aligned with reference images. We also display additionally generated tactile images in Fig.~\ref{fig:generated_results}. More tactile prediction results are provided in Fig.~\ref{fig:pred_results_more}.

\newpage

\newpage

\end{document}